\documentclass[final]{cvpr}

\usepackage{times}
\usepackage{epsfig}
\usepackage{graphicx}
\usepackage{amsmath}
\usepackage{amssymb}
\usepackage{comment}
\usepackage{color}
\usepackage{multirow}
\usepackage{paralist}
\usepackage{adjustbox}

\usepackage{bm}
\usepackage{amsthm,amsmath,amssymb}
\usepackage{mathrsfs}
 {
      \theoremstyle{plain}

  }
\usepackage{subcaption}
\usepackage[linesnumbered,ruled,vlined]{algorithm2e}
\usepackage{algorithmic}
\SetKwInput{KwInput}{Input}                
\SetKwInput{KwOutput}{Output}              

\usepackage[pagebackref=true,breaklinks=true,colorlinks,bookmarks=false]{hyperref}

\usepackage{color}

\def\cvprPaperID{7695} 
\def\confYear{CVPR 2021}

\usepackage{amsmath,amsfonts,bm}

\def\eqref#1{(\ref{#1})}

\def\1{\bm{1}}

\def\eps{{\epsilon}}

\def\rvb{{\mathbf{b}}}

\def\rvg{{\mathbf{g}}}

\def\rvx{{\mathbf{x}}}

\def\rmG{{\mathbf{G}}}

\def\rmW{{\mathbf{W}}}

\def\vzero{{\bm{0}}}

\DeclareMathAlphabet{\mathsfit}{\encodingdefault}{\sfdefault}{m}{sl}
\SetMathAlphabet{\mathsfit}{bold}{\encodingdefault}{\sfdefault}{bx}{n}

\def\gN{{\mathcal{N}}}

\begin{document}

\title{Robust Text CAPTCHAs Using Adversarial Examples}

\author{Rulin Shao\\
Xi'an Jiaotong University\\
{\tt\small shaorulin@stu.xjtu.edu.cn}
\and
Zhouxing Shi\\
University of California, Los Angeles\\
{\tt\small zhouxingshichn@gmail.com}
\and
Jinfeng Yi\\
JD AI Research\\
{\tt\small yijinfeng@jd.com}
\and
Pin-Yu Chen\\
IBM Research\\
{\tt\small pin-yu.chen@ibm.com}
\and
Cho-Jui Hsieh\\
University of California, Los Angeles\\
{\tt\small chohsieh@cs.ucla.edu}
}

\maketitle

\newcommand{\bl}[1]{{\color{blue} (#1)}}

\begin{abstract}
CAPTCHA (Completely Automated Public Truing test to tell Computers and Humans Apart) is a widely used technology to distinguish real users and automated users such as bots. However, the advance of AI technologies weakens many CAPTCHA tests and can induce security concerns. In this paper, we propose a user-friendly text-based CAPTCHA generation method named Robust Text CAPTCHA (RTC). At the first stage, the foregrounds and backgrounds are constructed with randomly sampled font and background images, which are then synthesized into identifiable pseudo adversarial CAPTCHAs. At the second stage, we design and apply a highly transferable adversarial attack for text CAPTCHAs to better obstruct CAPTCHA solvers. Our experiments cover comprehensive models including shallow models such as KNN, SVM and random forest, various deep neural networks and OCR models. Experiments show that our CAPTCHAs have a failure rate lower than one millionth in general and high usability. They are also robust against various defensive techniques that attackers may employ, including adversarial training, data pre-processing and manual tagging.
\end{abstract}

\begin{figure}[hbt!]
\centering
\includegraphics[width=.48\textwidth]{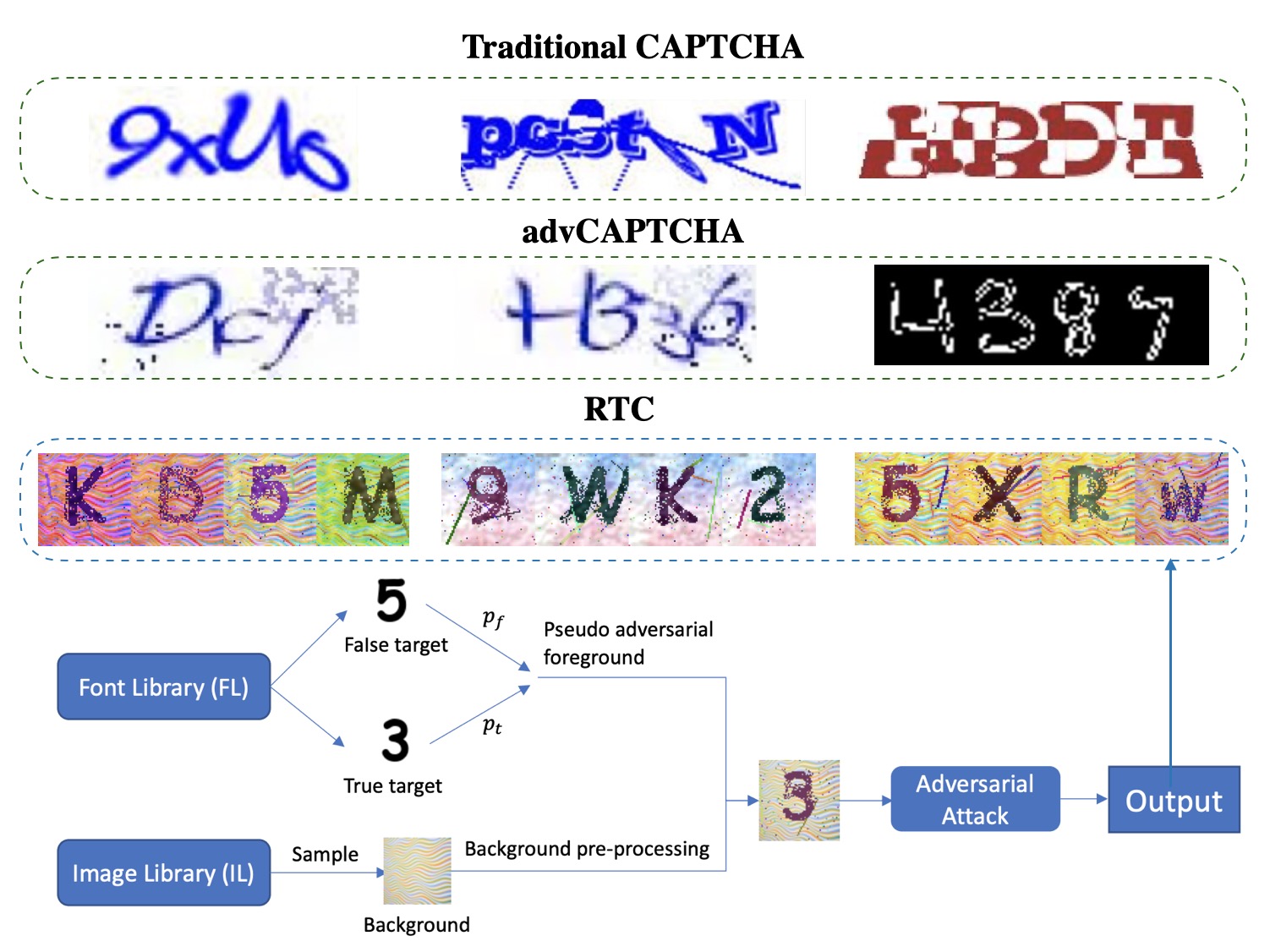}
\caption{Examples of different text-based CAPTCHAs and the method framework of our proposed robust text CAPTCHA (RTC) generation. The first row lists three traditional text-based CAPTCHAs generated by the BotDetect CAPTCHA Generator~({\footnotesize\url{https://captcha.com}}). The second row shows adversarial CAPTCHAs from prior works \cite{shi2019adversarial,shi2020text}. The third row presents some samples generated by our RTC method, which is more user-friendly and can be more diverse.}
\label{fig:captcha}
\end{figure}

\section{Introduction}
CAPTCHA (Completely Automated Public Truing Test to Tell Computers and Humans Apart) was originally designed to prevent bots or malwares from interacting with a web page~\cite{von2003captcha}. 
As the name indicates, CAPTCHA aims at identifying whether a user is a person or a bot using a specific test.
It has been widely used to prevent malicious automated registration, mass spamming and  bot programs.  

The most common form of CAPTCHA is an image of several distorted characters, which is called text-based CAPTCHA~\cite{von2008reCAPTCHA} consisting of random strings of letters and numbers.
Different approaches are utilized to distort characters and make them harder for bots to recognize. 
Some stretch and bend characters in ways as being looked through melted glass. Others put the characters behind a crosshatched pattern of bars to break up the shape of the characters, or use different colors or a field of dots~\cite{von2003captcha, singh2014survey, bursztein2011text}. 
In addition to text-based ones, there are also other categories of CAPTCHAs: 
Image-based CAPTCHAs \cite{datta2005imagination, gao2010novel, vikram2011semage} release an image selection task each time, and audio CAPTCHAs \cite{choudhary2013understanding, singh2014survey, meutzner2015constructing} present a series of spoken letters or numbers usually with distorted voice or background noise. In this paper, we focus on the most widely-used text-based CAPTCHAs. 

Despite the wide application of CAPTCHAs, as CAPTCHAs are reinforced with distortions to be harder for bots to recognize, they are also becoming increasingly difficult for humans. In the ``traditional CAPTCHA'' examples shown in Figure \ref{fig:captcha}, distortions can make the CAPTCHAs illegible for humans while still easy for bots to bypass \cite{baecher2011breaking}. This goes against the original intention of the CAPTCHA design. Besides, other traditional CAPTCHAs could be rather weak when attackers collect some examples and train a classification model as the CAPTCHA solver. Polakis~\cite{sivakorn2016m} solved Google's image CAPTCHAs with 70 percent accuracy, and Bock~\cite{bock2017unCAPTCHA} achieved more than 85\% accuracy on bypassing Google's audio CAPTCHA challenges~\cite{sano2013solving, tam2009breaking} using Google's own audio recognition service.

On the other hand, despite the high capacity of deep neural networks (DNN), research shows DNNs are vulnerable to small input perturbations~\cite{szegedy2013intriguing, goodfellow2014explaining}, which casts light on CAPTCHA design. To defend against deep neural network CAPTCHA solvers, some works constructed adversarial examples to fool deep models~\cite{kwon2020robust, shi2019adversarial, shi2020text}. However, \cite{kwon2020robust} applies white-box attacks and requires access to CAPTCHA solvers of attackers, which is an impractical assumption instead of making the generation solver-agnostic.
\cite{shi2019adversarial, shi2020text} propose to apply black-box attack to MNIST letters \cite{deng2012mnist} or existing CAPTCHA images, but \cite{shi2020text} still have a high recognition rate by CAPTCHA solvers (higher than 30\% as they show in the paper) and rely on queries to the inferred attack models to adapt their attacking strategies. Moreover, none of them considered a more practical situation when attackers collect a training dataset from deployed CAPTCHAs with human annotation, to train a stronger CAPTCHA solver. 

In this paper, we propose a user-friendly text-based CAPTCHA generation method, named \emph{Robust Text CAPTCHA (RTC)}.
We design pseudo adversarial foregrounds and backgrounds respectively and implement a scaled Gaussian translation with channel shifts attack (SGTCS) to strengthen the fidelity of the generated CAPTCHAs. Moreover, our RCT method demonstrates much lower recognition rates by CAPTCHA solvers compared to prior works while attaining high usability, making the resulting CAPTCHAs easy for human but hard for AI bots, as presented in Section \ref{sec:exp}.
Our main contributions are summarized as follows:
\begin{compactitem}
    \item We propose to generate effective text CAPTCHAs with a novel design of pseudo adversarial foreground and background images respectively. This design allows flexibility and diversity to the generation and makes the CAPTCHAs more robust in practice while retaining usability.
    \item We design a highly transferable adversarial attack for generating text CAPTCHAs,
    which focuses on increasing the transferability while preserving the readability of the foreground and adding diversity to the CAPTCHAs, instead of achieving a minimum disturbance. We also consider the impact brought by the low dimensional feature space of the text CAPTCHAs, and propose a transformation based attack to further promote transferability.
    \item Our experiments cover comprehensive CAPTCHA solver models including those based on deep neural networks, and also shallow models (KNN, SVM and random forest) and advanced OCR techniques.     
    Experiments show that our CAPTCHAs have a recognition rate by CAPTCHA solvers below one millionth and also high usability.
    Our method is also shown to be robust against many defensive techniques that may be utilized in CAPTCHA solvers.
\end{compactitem}

\section{Background}
In this section, we briefly introduce adversarial examples and defense strategies.

\subsection{Adversarial Attack}
Recent studies have shown that DNNs are vulnerable to adversarial examples. Imperceptible perturbation could fool the networks to make the wrong decisions~\cite{goodfellow2014explaining,szegedy2013intriguing}. According to whether the attacker could get access to the model, the adversarial attack could be broadly divided into white-box attack and black-box attack. In the case of CAPTCHA design, the potential CAPTCHA solvers are agnostic to the designer, and thus conducting black-box attack appears to be more reasonable. The black-box attack is commonly further categorized into transfer-based and query-based. The transfer-based black-box attack takes use of the transferability of adversarial examples. It usually attacks a surrogate model using a white-box attack, and the generated adversarial examples often transfer to other models. The query-based adversarial attack updates the examples utilizing the feedback of the model, which is more complex and time-consuming. It also concerns the designer to analyze the security risk of users to be bots before sending queries to get usable information. Thus the transfer-based adversarial attack is more practical when considering the CAPTCHA design. 

Adversarial attack can be formulated as a constrained optimization problem as
\begin{equation}
\underset{\rvx^{adv}}{\arg \max }~ J\left(\rvx^{adv}, y\right) \quad \text { s.t. }\left\|\rvx^{adv}-\rvx^{real}\right\|_{\infty} \leq \epsilon, 
\label{eq:attack}
\end{equation}
where the goal is to maximize the loss function of the classifier, denoted as $J\left(\rvx^{adv}, y\right)$, under the constraint that the adversarial example should be sufficiently close to the original example. 
White-box attacks usually calculate the gradient of the loss function with respect to the input for (approximately) solving \eqref{eq:attack}. Transfer-based black-box attack chooses a surrogate model for gradient calculation and then transfer to other models in a black-box manner. We summarize several white-box attack algorithms below:

\vspace{-0.3cm}
\paragraph{Fast Gradient Sign Method (FGSM)~\cite{goodfellow2014explaining}} is a one-step gradient ascent method that updates the input in a linear direction:
\begin{equation}
\rvx^{adv}=\rvx^{r e a l}+\epsilon \cdot \operatorname{sign}\left(\nabla_{\rvx} J\left(\rvx^{r e a l}, y\right)\right)
\end{equation}
where $J$ is the objective function of the classifier, $\epsilon$ is the attack radius, and $\rvx^{real}$, $\rvx^{adv}$, $y$ are the input image, adversarial example and the corresponding label respectively.

\vspace{-0.3cm}
\paragraph{Iterative Fast Gradient Sign Method (I-FGSM)}~\cite{kurakin2016adversarial} extends FGSM by iteratively calculating the gradient ascent direction and updating the input:
\begin{equation}
\rvx_{t+1}^{a d v}=\rvx_{t}^{a d v}+\alpha \cdot \operatorname{sign}\left(\nabla_{\rvx} J\left(\rvx_{t}^{a d v}, y\right)\right),
\end{equation}
where $t$ stands for the $t$-th iteration. The other calculations remain the same as the FGSM.

\vspace{-0.3cm}
\paragraph{Momentum Iterative Fast Gradient Sign Method (MI-FGSM)}\cite{dong2018boosting} further utilizes the past updating direction as a momentum term to improve the transferability:
\begin{equation}
\begin{array}{c}\rvg_{t+1}=\mu \cdot \rvg_{t}+\frac{\nabla_{\rvx} J\left(\rvx_{t}^{a d v}, y\right)}{\left\|\nabla_{\rvx} J\left(\rvx_{t}^{a d v}, y\right)\right\|_{1}}, \\ \rvx_{t+1}^{a d v}=\rvx_{t}^{a d v}+\alpha \cdot \operatorname{sign}\left(\rvg_{t+1}\right),\end{array}
\end{equation}
where $\mu$ is the momentum factor, $\bm{g_0}=0$ and $\bm{g_t}$ is the gradient direction for the past $t$ iterations.
\subsection{Defense Strategies}
Some defense methods are proposed to address the security concerns. And such methods may be utilized by attackers to strengthen their CAPTCHA solvers against adversarial CAPTCHAs, with examples introduced below:

\vspace{-0.3cm}
\paragraph{Adversarial Training}~\cite{kurakin2016adversarial}  incorporates adversarial examples in the training set to make the model robust to such perturbations. It demonstrate good performance against the specific attack method but is weak to the unseen ones \cite{stutz2020confidence}.

\vspace{-0.3cm}
\paragraph{Data Preprocessing}~\cite{buckman2018thermometer, guo2017countering} do not change the structure of the model and only performs transformation before feeding the inputs into the unmodified models in order to mitigate adversarial perturbation.
\vspace{-0.3cm}

\paragraph{Manual Labeling}
In addition, it is also reasonable to consider that attackers using CAPTCHA solvers may collect training examples right from deployed CAPTCHAs and employ human annotators to manually label them. With such training examples, they tend to obtain stronger CAPTCHA solvers for breaking CAPTCHAs, which has not been addressed in prior works. In Sec.~\ref{sec:manual}, we demonstrate the robustness of our RTC against such CAPTCHA solvers.

\section{RCT: Proposed Method}
Our adaptive adversarial CAPTCHA generation process could be divided into two parts. Firstly, we sample textured images and fonts randomly from font and image libraries respectively, and generate pseudo adversarial foregrounds and pre-processed backgrounds separately. Then the foregrounds and backgrounds are composited as the basic images which will be fed into the adversarial attack module to obtain the final images.

In the attack stage, we propose a scaled Gaussian translation with channel shifts attack (SGTCS) for adversarial text CAPTCHA generation. We utilize three techniques to boost the transferability of adversarial examples: weighted spatial translation, image scaling and channel shifts. These techniques could serve as data augmentation to help the attack method escape from the local maximum.

\subsection{Pseudo Adversarial Foreground and Diverse Background Generation}
We propose to design foreground and background images respectively and add simple but effective pseudo adversarial noises to the original images. The generated images are prepared to be further fed into the adversarial attack module, and are more user-friendly and diverse compared with the distorted traditional CAPTCHAs.\\

\vspace{-0.2cm}
\noindent\textbf{Foreground}
As shown in Figure \ref{fig:captcha}, for the foreground, we first sample a font from the font library (FL), which is not limited to digital fonts but also potentially handwritten ones from datasets such as MNIST~\cite{deng2012mnist} and EMNIST~\cite{cohen2017emnist}.
We then generate two characters, one with the true target label and one with a misleading label. The foreground is a composite of pixels sampled from the true and false characters with the possibility of $p_t$ and $p_f$ respectively. 
We take $p_t=0.9$ and $p_f=0.4$ by default in the consideration of maintaining a balance between improving human recognition and hindering machine recognition. 
The visual confusion of the true character with the wrong one is consistent with the common phenomenon that appeared when applying adversarial attacks to some low-dimensional feature space like MNIST~\cite{borji2019white}, and thus we name it as a \emph{pseudo adversarial foreground}.

\vspace{-0.4cm}
\paragraph{Background}
The background image is designed independently of the foreground in order to enlarge generation diversity. It is sampled from a library of images collected from the internet, and optionally it can also be made more challenging by applying image processing techniques, such as rotation, blurring, erosion, etc. 
The choice of the background image and pre-processing techniques are both configurable, which makes the generated images more diverse and thereby robust to various practical situations. 

\subsection{Scaled Gaussian Translation with Channel Shifts Attack}

To improve the robustness against CAPTCHA solvers, we utilize transfer-based adversarial attacks to generate adversarial CAPTCHAs and focus on improving the transferability. 
Prior research has shown that the convolutional neural networks have invariant properties when conducting parallel shifts~\cite{dong2019evading} and scaling~\cite{lin2019nesterov}. Based on these phenomena, some works have demonstrated that introducing image transformations to the inputs could improve the transferability of  attacks~\cite{xie2019improving}. We follow this path and incorporate Gaussian weighted spatial translation and image scaling to the attack process to further improve the transferability. However, different from the traditional attacks that aim at fooling classifiers with imperceptible perturbations, we incorporate channel shifts with an amplitude obeying uniform distribution under the constrain of the maximum disturbance. This could add diversity to the final CAPTCHA images while preserving the readability for humans.

We propose the following objective for our adversarial attack: 
\begin{equation}
\label{eq:objective}
    \underset{\rvx^{adv}}{\arg\max} \quad\widetilde{J}(\rvx^{adv})\qquad
    \text{s.t.} \quad \|\rvx^{adv}-\rvx\|_{\infty}\leq \epsilon,
\end{equation}
$$\text {where} \,\widetilde{J} (\rvx^{adv}) = \sum_{k=0}^m\alpha_k\sum_{i,j}\frac{e^{-\frac{i^2+j^2}{2\sigma^2}}}{2\pi \sigma^2}J(\mathcal{T}_{ij}(S_k(x^{adv})), y),$$
$J(\cdot)$ is the loss function for the classification task, $\mathcal{T}_{ij}$ stands for parallel shifts by $i$ and $j$ pixels respectively along the two axes of the image and $S_k$ is a scaling factor, weighted by $\alpha_k$. $\sum_k\alpha_k=1$ and we choose $\alpha_k=\frac{1}{m}$ by default. $\rvx^{adv}$ and $\rvx$ stand for the adversarial example and the original image respectively, and $\epsilon$ is a pre-defined perturbation radius. 
Our goal is to find the perturbation such that the resulting adversarial example will maximize the average loss under spatial transformations including scaling and shifts.

We utilize Nesterov accelerated gradient (NAG)~\cite{nesterov1983method, lin2019nesterov, dozat2016incorporating} with gradient-based iterative attack~\cite{kurakin2016adversarial, madry2017towards} to optimize our adversarial perturbation. Similar to the momentum method~\cite{sutskever2013importance, dong2018boosting}, NAG looks forward before updating:
\begin{align}
    \rvx_t^{nes} &= \rvx_t^{adv} + \alpha \cdot \mu \cdot \bm{g_t},\\
    \rvg_{t+1} &= \mu \cdot \rvg_t + \frac{\bigtriangledown_{\rvx}\widetilde{J}(\rvx_t^{nes}, y^{true})}{\|\bigtriangledown_{\rvx}\widetilde{J}(\rvx_t^{nes}, y^{true})\|_1},
\end{align}
where $\mu$ is a momentum factor that controls the influence of the previous gradient on the next gradient update. 
The adversarial perturbation is then updated by
\begin{equation}
    \rvx_{t+1}^{adv}=Clip_{\rvx,\eps}(\rvx_t^{adv}+\alpha \cdot sign(\bm{g_{t+1})}),
\end{equation}
where $Clip^{\epsilon}_{\rvx,\eps}(\cdot)$ stands for clipping the new example to satisfy $\| \rvx_{t+1}^{adv}-\rvx\|_\infty\leq\eps$.

However, the computation of $\tilde{J}$ is complex because it needs to feed the images into the neural network and calculate the corresponding gradients respectively and repeatedly for each kind of spatial transformations. Therefore, we sought to find a more efficient way to calculate the gradients.
Following \cite{dong2019evading}, we assume the gradients of the objective function are approximately the same as the original ones after conducting small parallel shifts and scaling to the inputs \cite{lin2019nesterov}. Then we can derive a simpler form for the gradient computation of Eq. (\ref{eq:objective}):

\small
\begin{equation}\label{eq:deri}
\begin{aligned}
    &\bigtriangledown_{\rvx}\left(\sum_{k=0}^m\alpha_k\sum_{i,j}\frac{e^{-\frac{i^2+j^2}{2\sigma^2}}}{2\pi \sigma^2}J(\mathcal{T}_{ij}(S_k(\rvx)), y)\right)|_{\rvx=\bm{\hat{x}}}\\
    &\approx \sum_{k=0}^m \alpha_k  \sum_{i,j}\frac{e^{-\frac{i^2+j^2}{2\sigma^2}}}{2\pi \sigma^2}\mathcal{T}_{-i-j}(\bigtriangledown_{\rvx}J(\rvx,y)|_{\rvx=\bm{\hat{x}}}) \\
    &= \sum_{i,j}\frac{e^{-\frac{i^2+j^2}{2\sigma^2}}}{2\pi \sigma^2}\mathcal{T}_{-i-j}( \sum_{k=0}^m \alpha_k (\bigtriangledown_{\rvx}J(\rvx,y)|_{\rvx=\bm{\hat{x}}})) \\
    &= \rmG* (\sum_{k=0}^m \alpha_k \bigtriangledown_{\rvx}J(\rvx,y)|_{\rvx=\bm{\hat{x}}}),
\end{aligned}
\end{equation}\normalsize
which can be achieved by convolving the image with a two dimensional Gaussian kernel $\rmG$ defined as $G_{ij}= \frac{e^{-\frac{i^2+j^2}{2\sigma^2}}}{2\pi \sigma^2}$. 

Besides, we also conduct stochastic channel shifts to achieve better diversity. The degree of the shifts obeys a uniform distribution $b \sim \mathcal{U}(0, \epsilon)$, where $\epsilon$ is the maximum attack radius. This can be achieved by applying a convolutional layer to post-process the gradient calculation. The weights of this layer $W_{G}$ conform to a two-dimensional Gaussian distribution $\mathcal{N}(\bm{0}, \bm{\sigma}^2)$, and the bias $b_{u}$ follows a uniform distribution $\mathcal{U}(0, \epsilon)$. The generation process is presented in Algorithm \ref{alg:RTC}.

Notice the right hand side (RHS) is a convolution multiplication. Taking the RHS of Eq.(\ref{eq:deri}) and applying a Fourier Transform on it, the convolution operator is converted into a dot product:
\begin{equation}\label{eq:dot}
    \mathcal{F}^{-1}(\sum_{k=0}^m  e^{\frac{-2\pi^2\rho^2}{\alpha_k^2}}\cdot \mathcal{F}(\bigtriangledown_{\rvx}J(\rvx,y))).
\end{equation}
From Eq.(\ref{eq:dot}), the frequency of the gradient goes through a Gaussian low-pass filter, preserving the perturbation on the response frequency\cite{xu2019frequency}. According to the F-Principle \cite{xu2019frequency}, it is a general phenomenon that DNNs fit target functions from low to high relative frequencies during the training. Thus applying a carefully crafted Gaussian low-pass filter to the first several updating iterations could have better generalization property in this sense. It is also consistent with the gradient smoothing~\cite{sharma2018caad} technique which is widely used to improve the transferability of adversarial examples.

\begin{algorithm}[tb]
\SetAlgoLined
\KwOutput{An adversarial CAPTCHA}
 Generate pseudo adversarial foreground $\rvx_f$ and background $\rvx_b$\;
 Composite $\rvx_f$ and $\rvx_b$ as $\rvx_{pseudo}$\;
 Initialize attack parameters $\alpha = \epsilon / T$, $\mu$, $\sigma$\; $\rvg_0=\vzero$\; $\rvx_0^{adv}=\rvx_{pseudo}$\;
 \For{$t=0$\ to\ $T-1$}
 {
 $\rvx_t^{nes} = \rvx_t^{adv} + \alpha \cdot \mu \cdot \rvg_t $\;
  $\hat{\rvg}_{t+1} = \mu \cdot \rvg_t + \frac{\bigtriangledown_{\rvx}\widetilde{J}(\rvx_t^{nes}, y^{true})}{\|\bigtriangledown_{\rvx}\widetilde{J}(\rvx_t^{nes}, y^{true})\|_1}$\;
  Sample $\rmW_G \sim \mathcal{N}(0, \bm{\sigma^2}), \rvb_u \sim \mathcal{U}(0, \epsilon)$\;
  $\Tilde{\rvg}_{t+1} = \rmW_G * \hat{\rvg}_{t+1} + \rvb_u$\;
  $\rvg_{t+1}=\mu\cdot\rvg_t + \frac{\Tilde{\rvg}_{t+1}}{\| \Tilde{\rvg}_{t+1} \|_1}$\;
  $\rvx_{t+1}^{adv}=Clip_{\rvx,\eps}(\rvx_t^{adv}+\alpha \cdot sign(\bm{g_{t+1})})$ \;
 }
$\rvx^{adv}=\rvx_T^{adv}$\;
\caption{Robust Text CAPTCHA (RTC) generation}
\label{alg:RTC}
\end{algorithm}

\section{Experiments}\label{sec:exp}

In our experiments, we demonstrate the feasibility and robustness of our method by comprehensively considering techniques that attackers may utilize in CAPTCHA solvers to bypass our CAPTCHAs.
We first evaluate the recognition rate of normally trained CAPTCHA solvers using examples without pseudo adversarial perturbations for training.
We then also try to enhance CAPTCHA solvers with adversarial training and image pre-processing filters.
Next, we also consider a scenario that the an attacker collect and manually label a number of our adversarial CAPTCHAs for training, and we test the recognition rate of CAPTCHA solvers on our unseen adversarial CAPTCHAs.
Moreover, we also have usability test involving human annotators to verify the easiness of our adversarial CAPTCHAs for human users.
And we also have an ablation study to show the effectiveness of each part of the adversarial CAPTCHA generation. Model analysis with regard to hyper-parameters are presented in the supplementary material.

\subsection{Settings}\label{sec:setting}

\paragraph{Basic Images Generation}
English letters and numbers are utilized to generate the basic text. Among the 52 letters and 10 numbers, we delete number '0' and leters 'C', 'c', 'O', 'o', 'S', 's' as they are essentially confusing to humans. We utilize 7 fonts to generate characters as foregrounds, and choose 9 simple images as the backgrounds. The fonts include Herculanum, Papyrus, AmericanTypewriter, MarkerFelt, Brush Scropt, Times New Roman Italic and Arial Italic, which all appear to be regular and can thus be easily recognized by humans. We resize the background images into $64\times 64$ pixels and draw characters on the background with a randomly sampled font. We also implement some post-processing techniques in common CAPTCHA generation process including adding noise, erosion, rotating, etc. The post-processing could boost the diversity of the basic images on which we will train CAPTCHA solver models and conduct adversarial attacks. Meanwhile, we keep the degree of the modification relatively small to maintain the easiness for humans.

\vspace{-0.2cm}
\paragraph{CAPTCHA Solvers}
The adversarially generated CAPTCHAs are supposed to defend against various kinds of CAPTCHA solvers, especially the powerful ones with deep neural networks. In our experiment, we consider three categories of CAPTCHA solvers: traditional feature engineering models, deep neural network based models and Optical Character Recognition (OCR) models. 
For the traditional feature engineering ones, we utilize K-Nearest Network (KNN)~\cite{altman1992introduction}, Support Vector Machine (SVM)~\cite{suykens1999least} and random forest classifer (RFC)~\cite{liaw2002classification, ho1995random}. Besides the vanilla versions of the aforementioned models, we also combine these models with different feature descriptors such as Histogram of Oriented Gradient (HOG)~\cite{dalal2005histograms}. By combining different traditional classification methods with different feature descriptors, we have 9 kinds of shallow models involved.
For the deep neural networks, we investigate 16 models: LeNet~\cite{lecun1998gradient}, AlexNet~\cite{krizhevsky2017imagenet}, VGG11, VGG13, VGG16, VGG19~\cite{simonyan2014very}, GoogLeNet~\cite{szegedy2015going}, ResNet18, ResNet34, ResNet50, ResNet101, ResNet152~\cite{he2016deep}, DenseNet121, DenseNet161, DenseNet169 and DenseNet201~\cite{huang2017densely}. 
For OCR models, we investigate two Software Development Kits (SDKs), Huawei OCR SDK\footnote{\url{https://ai.baidu.com/sdk\#ocr}} and Baidu OCR SDK\footnote{\url{https://www.huaweicloud.com/product/ocr/image-to-txt.html}} respectively. We set the language type option to English for Baidu OCR SDK to reduce the difficulty of recognition. The result of OCR recognition must be exactly the same as the label to be considered as correct. Additions and deletions of letters and recognition errors will be regarded as recognition failure.
Due to space limitations, we only present the result of some representative models, and the rest experiments can be found in the supplementary material.

\vspace{-0.2cm}
\paragraph{Adversarial Attacks}
For adversarial attacks, we choose the surrogate model from VGG19 and ResNet50 by default, as we find adversarial examples generated on these two models show better transferability to other models. The transferability test can be found in the supplemental material. The perturbation radius $\epsilon$ is set to $0.1$ and the number of attack iterations is set to $10$. And we set the momentum factor $\mu$ to 1 which shows the best performance in  prior works~\cite{dong2018boosting}. The default $\mu$ and $\sigma$ for the two-dimensional normal distribution is set to $0$ and $3$.

\subsection{Defend Against Normally Trained Models}\label{sec:normal}

We generate some training CAPTCHAs to train CAPTCHA solvers.
In this section, we consider normally trained models, and thus the training CAPTCHAs are generated without going through the pseudo adversarial foreground generation process -- these are relatively considered as clean examples. 
Other settings for generation remain the same as we directly consider the situation when CAPTCHA solvers could collect the same data as ours.
The CAPTCHA solver models are trained with 2750 examples for 200 epochs, achieving an accuracy above 80\% for DNNs and a maximum at 66.72\% for shallow models. We also utilize Baidu and Huawei OCR SDKs and classify them as normaly trained ones.

\begin{table}[t!]
  \begin{center}
    \caption{Recognition rate by normally trained  CAPTCHA solver models on test CAPTCHAs with different lengths. ``RTC'' stands for our adversarial CAPTCHAs with the pseudo adversarial foreground generation process. ``Clean'' does not have this process and its examples are similar as those used for training. The surrogate model is denoted with ``*''.}
    \label{tab:dnn}
    \resizebox{\columnwidth}{!}{
    \begin{tabular}{l|cc|cc|cc}
      \hline
      \textbf{Length} & \multicolumn{2}{c|}{\textbf{1 digit}} & \multicolumn{2}{c|}{\textbf{4 digits}} & \multicolumn{2}{c}{\textbf{6 digits}} \\
      \hline
      \textbf{CAPTCHA} & \textbf{RTC} & \textbf{Clean} & \textbf{RTC} & \textbf{Clean} & \textbf{RTC} & \textbf{Clean} \\
      \hline
      VGG19* & 0.00 & 0.83 & 0.00e+00 & 0.46 & 0.00e+00 & 0.32\\
      \hline
      LeNet & 0.03 & 0.88 & 7.57e-07 & 0.60 & 6.59e-10 & 0.47\\AlexNet & 0.07 & 0.89 & 2.41e-05 & 0.61 & 1.19e-07 & 0.48\\GoogLeNet & 0.02 & 0.99 & 2.39e-07 & 0.95 & 1.17e-10 & 0.93\\ResNet50 & 0.03 & 0.98 & 1.21e-06 & 0.92 & 1.34e-09 & 0.89\\DenseNet169 & 0.06 & 0.97 & 1.55e-05 & 0.87 & 6.08e-08 & 0.81\\RFC & 0.01 & 0.45 & 4.67e-08 & 0.04 & 1.01e-11 & 0.01\\KNN+HOG & 0.06 & 0.57 & 1.21e-05 & 0.11 & 4.22e-08 & 0.04\\SVM+HOG & 0.06 & 0.67 & 1.21e-05 & 0.20 & 4.22e-08 & 0.09\\
      Baidu OCR & 0.00 & 0.19 & 0.00 & 1.30e-03 & 0.00 & 4.70e-05 \\
      Huawei OCR & 0.03 & 0.04 & 1.21e-06 & 2.56e-06 & 1.34e-09 & 4.10e-09\\
      \hline
    \end{tabular}
    }
  \end{center}
\end{table}

We evaluate the recognition rate of these models on test CAPTCHAs, and we show the results of some models in Table~\ref{tab:dnn}, while the results of other models with similar structures (e.g., VGG11, VGG13 and VGG16 are similar to VGG19) in the supplementary material. 
Table~\ref{tab:dnn} reflects the higher capacity of DNNs in recognition work compared with the shallow ones. OCRs tend to perform better when the captcha outline is clear and standard and the background is simple. But they exhibit limited performance against our design especially when we choose relatively complex images as the background. It is worth noting that our RTC has reached a recognition rate less than one sample per million for all these models.

\subsection{Defend Against Adversarially Trained Models}

As attackers who train CAPTCHA solvers may utilize adversarial training to make the solvers more robust to potential adversarial CAPTCHAs, we also evaluate the accuracy of adversarially trained models on our adversarial CAPTCHAs. We implement the adversarial training following the settings in \cite{kurakin2016adversarial}. The loss function for adversarial training is a weighted combination of the losses computed by clean images and adversarially generated images:
\begin{equation}
\resizebox{\columnwidth}{!}{%
$Loss=\frac{1}{(m-k)+\lambda k}\left(\sum_{i=0}^{m-k-1} J\left(X_{i} \mid y_{i}\right)+\lambda \sum_{i=0}^{k-1} J\left(X_{i}^{a d v} \mid y_{i}\right)\right),$%
}
\end{equation}
where $\lambda\!=\!0.3$, $k\!=\!16$, $m\!=\!32$ and $m$ is the batch size. 
Besides, the perturbation amplitude to generate the adversarial counterpart is chosen randomly and independently, which is drawn from a truncated normal distribution with underlying $\gN\left(\mu=0, \sigma=8\right)$ truncated in interval $\left[0, 16\right]$. This helps the adversarially trained model become robust to a wide range of $\epsilon$. We conduct adversarial training with AlexNet, GoogLeNet, VGG19, ResNet50 and DenseNet169 and the results are shown in Table \ref{tab:adv_train}. The test recognition rate by adversarially trained mdoels is significantly lower than their training accuracy, e.g., dropped from $96.18\%$ to $3.96\%$ for ResNet50 and from $89.82\%$ to $2.21\%$ for VGG19. It demonstrates that our RTC is also robust to adversarially trained models.

\begin{table}[t!]
  \begin{center}
    \caption{Training accuracy (``train'') and test recognition rate (``test'') by adversarially trained models.} 
    \label{tab:adv_train}
    \resizebox{\columnwidth}{!}{
    \begin{tabular}{lcccccc}
      \hline
      & \textbf{ResNet50} & \textbf{VGG19} & \textbf{DenseNet169} & \textbf{GoogLeNet} & \textbf{AlexNet}\\
      \hline
      Train & 0.9618 & 0.8982 & 0.9036 & 0.9818 & 0.8982\\Test & 0.0369 & 0.0221 & 0.0443 & 0.0332 & 0.1771\\
      \hline
    \end{tabular}}
	  \end{center}
\end{table}

The work in \cite{kurakin2016adversarial} utilizes FGSM with random perturbation magnitudes, and demonstrates better transferability than iterative ones. However, our RTC integrates pseudo foreground generation and channel shifts into the process, which is different in nature to the traditional gradient-based attacks and thus neutralizes the effect of adversarial training.

\begin{table*}[t!]
  \begin{center}
  \small
    \caption{Recognition rate by models with image preprocessing filters.}
    \label{tab:preprocess}
    \resizebox{1.95\columnwidth}{!}{
    \begin{tabular}{lcccccc}
      \hline
      Filter \ \ \ \ & \ \ LeNet\ \  & \ \ AlexNet\ \  & \ \ GoogLeNet \ \ & \ \ VGG19\ \  & \ \ ResNet50 \ \  & \ \ DenseNet169 \ \ \\
      \hline
      None & 0.0295 & 0.0701 & 0.0221 & 0.0000 & 0.0332 & 0.0627 \\ BLUR & 0.0369 & 0.1107 & 0.1033 & 0.0332 & 0.0627 & 0.1697 \\ GaussianBlur & 0.0258 & 0.107 & 0.0590 & 0.0369 & 0.0812 & 0.1771 \\ DETAIL & 0.0295 & 0.0701 & 0.0258 & 0.0037 & 0.0258 & 0.0443 \\ SMOOTH & 0.0295 & 0.0923 & 0.0185 & 0.0037 & 0.0295 & 0.0738 \\ SHARPEN & 0.0295 & 0.0738 & 0.0185 & 0.0037 & 0.0258 & 0.0443 \\ SMOOTH MORE & 0.0258 & 0.0886 & 0.0185 & 0.0037 & 0.0295 & 0.0701 \\ FIND EDGES & 0.0295 & 0.0185 & 0.0185 & 0.0185 & 0.0185 & 0.0185 \\ EDGE ENHANCE & 0.0406 & 0.0849 & 0.0185 & 0.0037 & 0.0295 & 0.0332 \\ EDGE ENHANCE MORE & 0.0406 & 0.0923 & 0.0185 & 0.0037 & 0.0185 & 0.0295 \\ EMBOSS & 0.0258 & 0.0295 & 0.0185 & 0.0111 & 0.0185 & 0.0185 \\ CONTOUR & 0.0443 & 0.0517 & 0.0185 & 0.0258 & 0.0221 & 0.0221 \\ MinFilter & 0.0295 & 0.0221 & 0.0221 & 0.0074 & 0.0185 & 0.0185 \\ MaxFilter & 0.0406 & 0.0849 & 0.0148 & 0.0443 & 0.0332 & 0.0627 \\ MedianFilter & 0.0443 & 0.1033 & 0.0258 & 0.0111 & 0.0480 & 0.0849 \\ ModeFilter & 0.0295 & 0.0738 & 0.0258 & 0.0074 & 0.0221 & 0.0590 \\
      \hline
    \end{tabular}
    }
  \end{center}
\end{table*}

\subsection{Defend Against Image Processing Filters}

There are some \emph{image processing filters} which can be potentially used for denoising and somewhat reducing the effect of adversarial perturbations, as demonstrated in \cite{shi2019adversarial}.
Thus we also study the robustness of our CAPTCHAs against CAPTCHA solver models with such filters.
Specifically, we consider 15 filters available in the \texttt{ImageFilter} module of Pillow package\footnote{\url{https://pillow.readthedocs.io}}, such as \texttt{MedianFilter} which stands for picking the median pixel value in each filter window.
In this setting, input images firstly pass an image filter before being taken as the input of CAPTCHA solvers. We present results in Table \ref{tab:preprocess}.
Some filters such as \texttt{GaussianBlur} can help increase the recognition rate by a small degree, while some filters such as \texttt{FIND\_EDGES} even decreases the recognition rate.
Overall, the recognition rate by the CAPTCHA solvers remains low even when they apply image preprocessing filters, which further demonstrates the robustness of our RTC.

\subsection{Defend Against Manual Labelling}\label{sec:manual}

When a CAPTCHA system is deployed online, the generated CAPTCHAs are supposed to be open to users and also potential attackers.
Therefore, an attacker may collect a number of our generated CAPTCHAs, employ workers to manually label them, and utilize these data to enhance their CAPTCHA solvers.
It is thus important to investigate whether the generated CAPTCHAs are challenging and costly enough for attackers to train new CAPTCHA solvers. Ideally, we hope the CAPTCHA solver models trained with a limited number of RTC examples can only achieve a low accuracy on unseen RTC examples.

We sample 1,000 RTC generated under the default setting described in Section \ref{sec:setting} and use them to train new classifiers for 200 iterations. 
Then we change the background library and introduce new images as the new background. Table \ref{tab:manual} shows the recognition rate of the new adversarial CAPTCHA character generated with 4 new background images.
The results  demonstrate that our approach leads to low test recognition rate when we use different backgrounds for test.
This demonstrates that it is costly and difficult for attackers to manually label data and train CAPTCHA solvers, while we can defense such attackers by simply changing the background images in generating new CAPTCHAs.

\begin{table*}[t!]
  \begin{center}
   \footnotesize
    \caption{The training accuracy, and test recognition rate per character on newly generated RTC images with different backgrounds (BG $0\sim 3$), when CAPTCHA solvers are trained with manually labeled RTC images. ``Avg test'' is short for average test recognition rate.}
    \label{tab:manual}
    \resizebox{1.8\columnwidth}{!}{
    \begin{tabular}{lcccccc}
      \hline
      \textbf{Solver}\ \ \ \ \ \  & \ \ \ \ \textbf{Training}\ \ \ \  &\ \ \ \ \textbf{BG 0}\ \ \ \  & \ \ \ \ \textbf{BG 1} \ \ \ \ &\ \ \ \  \textbf{BG 2}\ \  \ \ & \ \ \ \ \textbf{BG 3} \ \ \ \ &\ \ \ \  \textbf{Avg test}\ \ \ \  \\
      \hline
      LeNet & 1.0000 & 0.2156 & 0.0701 & 0.0183 & 0.0000 & 0.0723\\
      AlexNet & 1.0000 & 0.0186 & 0.0185 & 0.0183 & 0.0154 & 0.0176\\
      GoogLeNet & 1.0000 & 0.0149 & 0.0332 & 0.0147 & 0.0231 & 0.0241\\
      VGG19 & 0.0185 & 0.0186 & 0.0185 & 0.0183 & 0.0154 & 0.0176\\
      ResNet50 & 0.9815 & 0.1710 & 0.1107 & 0.0220 & 0.0385 & 0.1094\\
      DenseNet169 & 1.0000 & 0.0186 & 0.0185 & 0.0220 & 0.0154 & 0.0213\\
      \hline
    \end{tabular}
    }
  \end{center}
\end{table*}

\subsection{Usability Test}

To verify the usability of our generated adversarial CAPTCHAs, we conduct a usability test on Amazon Mechanical Turk (MTurk)\footnote{\url{https://www.mturk.com}}. Specifically, we sample 400 our adversarial CAPTCHAs consisting of four characters each that is generated under the default setting described in Section \ref{sec:setting}
, and present them to MTurk workers as simulated users of websites using our CAPTCHAs. The accuracy of these workers is 91.25\%, which is comparable to human accuracy on  ordinary CAPTCHAs reported in previous works, such as 89.8\% in \cite{shi2020text}. 
Besides, in terms of time cost, the average time the MTurk workers spend on each CAPTCHA is 8.5s,
which is also comparable to 7.5s on ordinary (non-adversarial) CAPTCHAs reported in \cite{shi2020text}. 
This verifies the usability of our CAPTCHAs -- they are similarly easy for human users compared to ordinary CAPTCHAs while they have the ability to defend against automatic CAPTCHA solvers.
In particular, we find that character ``l'' is the most error-prone for human, with a lowest accuracy of 81.25\%, and in $2/3$ of the error cases they are identified as ``i''. This is reasonable as ``l'' indeed appears to be close to ``i'', and in real deployment, such characters confusing for humans can be avoided in generation to further improve usability.

\subsection{Ablation Study}\label{sec:ablation}
We conduct an ablation study to show how each different techniques are contributing to the performance of RTC. We divide the whole process of RTC into two stages: Pseudo Adversarial Example Generation (PAEG) and Scaled Gaussian Translation with Channel Shifts (SGTCS). The PAEG consists of the basic foreground and background design and SGTCS is the adversarial attack we utilize to generate adversarial CAPTCHAs. We ablate each part respectively to compare the contribution of these two parts.

\begin{table}[t!]
  \begin{center}
    \caption{Results of ablation study. ``Both" means skipping both stages and use data generated in the same way as the normal training set. ``SGTCS" and ``PAEG" means ablating the adversarial attack and skipping pseudo adversarial foreground generation process respectively. ``None" is the situation we apply in the full process of RTC. The surrogate model is denoted with ``*''.}
    \label{tab:ablation}
    \begin{tabular}{lcccc}
      \hline
      \textbf{Solver} & \textbf{Both} & \textbf{SGTCS} & \textbf{PAEG} & \textbf{None}\\
      \hline
      ResNet50* & 0.9800 & 0.0024 & 0.0000 & 0.0000\\
      \hline
      LeNet & 0.8818 & 0.3269 & 0.0784 & 0.0480\\AlexNet & 0.8855 & 0.2663 & 0.1608 & 0.0812\\GoogLeNet & 0.9873 & 0.1985 & 0.0157 & 0.0185\\VGG19 & 0.8255 & 0.3051 & 0.0549 & 0.0480\\DenseNet169 & 0.9655 & 0.1937 & 0.0549 & 0.0480\\
      \hline
    \end{tabular}
  \end{center}
\end{table}

From Table \ref{tab:ablation}, SGTCS adversarial attack module contributes more to the low recognition rate by the CAPTCHA solvers compared with the PAEG module. However, PAEG also demonstrates a further reduction in the recognition rate. Besides, it makes the CAPTCHA generation more diverse and thus make them robust to a wide range of situations such as attackers using human annotation as presented in Section \ref{sec:manual}. More detailed on ablation study in terms of different translations, e.g. scaling and channel shifts, can be found in the supplementary material.

\subsection{Comparison with Baselines}

We reproduce the baseline methods in \cite{kwon2020robust} for comparison. We investigate 3 adversarial attack methods involved in their work: FGSM \cite{goodfellow2014explaining}, I-FGSM \cite{kurakin2016adversarial}, and MI-FGSM \cite{dong2018boosting}. The setting for this experiment is the same as the default one except that we choose ResNet50 as the surrogate model this time. The recognition rate by normally trained deep neural network is presented in Table \ref{tab:baseline}.
Our method demonstrates a much lower recognition rate on CAPTCHA solver models except the surrogate model, while the recognition rate of the surrogate model is 0 for all methods.
The results show that our RTC has better transferability and thus more practically effective compared with the baselines.

\begin{table}[t!]
  \begin{center}
    \caption{Recognition rate in comparison with the baseline methods. The surrogate model is denoted with ``*''.}
    \label{tab:baseline}
    \resizebox{\columnwidth}{!}{
    \begin{tabular}{lccccc} 
      \hline
      \multirow{2}{*}{\textbf{Solver}} & \multicolumn{3}{c}{Baseline by \cite{kwon2020robust}} & \multirow{2}{*}{\textbf{Ours}}\\
      & FGSM & I-FGSM & MI-FGSM & \\
      \hline
      ResNet50* & 0.00 & 0.00 & 0.00 & 0.00 \\
      \hline
      LeNet & 0.61 & 0.68 & 0.62 & 0.05\\
      AlexNet & 0.61 & 0.70 & 0.67 & 0.08\\
      GoogLeNet & 0.18 & 0.41 & 0.34 & 0.02\\
      VGG19 & 0.57 & 0.59 & 0.57 & 0.05\\
      DenseNet169 & 0.42 & 0.70 & 0.63 & 0.05\\
      \hline
    \end{tabular}
    }
  \end{center}
\end{table}

\section{Conclusion}
In this paper, we propose a user-friendly text-based CAPTCHA generation method, Robust Text CAPTCHA (RTC), for practical use. Our experiments comprehensively cover various CAPTCHA solver models including shallow learning models, deep neural networks, and optical character recognition (OCR) systems. The results demonstrate the effectiveness and robustness of RTC against CAPTCHA solvers with different training settings and defensive techniques.
We also consider an extreme situation when the CAPTCHA solvers employ manual labeling to utilize our RTC examples for training, while their recognition rate is still low on new test RTCs with changed backgrounds.
Our RTC shows lower recognition rate compared to baseline methods and also achieves high usability.

{\small
\bibliographystyle{ieee_fullname}

}


\end{document}